\documentclass[conference]{IEEEtran}
\IEEEoverridecommandlockouts
\usepackage{cite}
\usepackage{amsmath,amssymb,amsfonts}
\usepackage{algorithmic}
\usepackage{graphicx}
\usepackage{textcomp}
\usepackage{xcolor}

\usepackage{flushend} 

\def\BibTeX{{\rm B\kern-.05em{\sc i\kern-.025em b}\kern-.08em
    T\kern-.1667em\lower.7ex\hbox{E}\kern-.125emX}}
\begin{document}


\title{Discrete Fourier Transform-based Point Cloud Compression for Efficient SLAM in Featureless Terrain}

\author{
\IEEEauthorblockN{Riku Suzuki*}
\IEEEauthorblockA{\textit{Space Robotics Lab. (SRL),}\\ 
\textit{Department of Aerospace Engineering,}\\
\textit{Tohoku University}\\
Sendai, Japan \\
{\small \tt suzuki.riku.q1@dc.tohoku.ac.jp}}
*Corresponding author
\and
\IEEEauthorblockN{Ayumi Umemura}
\IEEEauthorblockA{\textit{Space Robotics Lab. (SRL),}\\
\textit{Department of Aerospace Engineering,} \\
\textit{Tohoku University}\\
Sendai, Japan \\
{\small \tt umemura.ayumi.t6@dc.tohoku.ac.jp}}
\and
\IEEEauthorblockN{Shreya Santra}
\IEEEauthorblockA{\textit{Space Robotics Lab. (SRL),}\\
\textit{Department of Aerospace Engineering,} \\
\textit{Tohoku University}\\
Sendai, Japan \\
{\small \tt shreya.santra@tohoku.ac.jp}}
\vspace{5mm}
\and
\IEEEauthorblockN{Kentaro Uno}
\hspace{95mm}
\IEEEauthorblockA{\textit{Space Robotics Lab. (SRL),}\\
\textit{Department of Aerospace Engineering,} \\
\textit{Tohoku University}\\
Sendai, Japan \\
{\small \tt unoken@tohoku.ac.jp}}
\and
\IEEEauthorblockN{Kazuya Yoshida}
\hspace{-150mm} 
\IEEEauthorblockA{\textit{Space Robotics Lab. (SRL),}\\
\textit{Department of Aerospace Engineering,} \\
\textit{Tohoku University}\\
Sendai, Japan \\
{\small \tt yoshida.astro@tohoku.ac.jp}}
}

\maketitle

\vspace{-2em}

\begin{abstract}

Simultaneous Localization and Mapping (SLAM) is an essential technology for the efficiency and reliability of unmanned robotic exploration missions. While the onboard computational capability and communication bandwidth are critically limited, the point cloud data handled by SLAM is large in size, attracting attention to data compression methods. To address such a problem, in this paper, we propose a new method for compressing point cloud maps by exploiting the Discrete Fourier Transform (DFT). The proposed technique converts the Digital Elevation Model (DEM) to the frequency-domain 2D image and omits its high-frequency components, focusing on the exploration of gradual terrains such as planets and deserts. Unlike terrains with detailed structures such as artificial environments, high-frequency components contribute little to the representation of gradual terrains. Thus, this method is effective in compressing data size without significant degradation of the point cloud. We evaluated the method in terms of compression rate and accuracy using camera sequences of two terrains with different elevation profiles.

\end{abstract}

\section{INTRODUCTION}
In the fields of robotics and autonomous driving, SLAM, a technique to simultaneously perform self-localization and map creation, has garnered widespread attention. Particularly in unmanned robotic missions in extreme environments such as lunar and planetary exploration, the robot's onboard calculation capacity and the operatable time period are severely limited. Thus, the accuracy and computational efficiency of such robots' SLAM directly affects the exploration coverage~\cite{Ke2019}, which is the most important index in the mission like in-situ resource discovery on the planetary surface~\cite{Ebadi2024}. 
\begin{figure}[t]
  \centering
  \includegraphics[width=.8\linewidth]{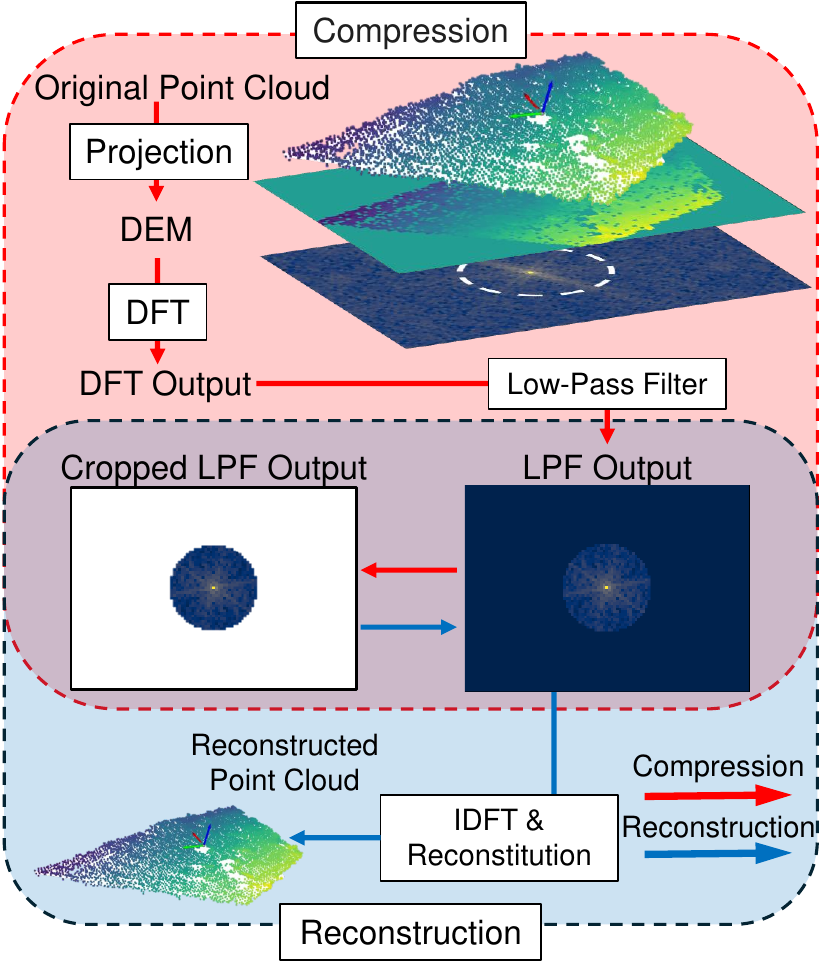} 
  \caption{Our technique employs Discrete Fourier Transform (DFT) to remove high-frequency components to compress the data size of the point cloud generated by SLAM. In the case of gradual terrains, the removal of high-frequency components has minimal impact, enabling the minimum error in the reconstructed point cloud after the reduction.
} %
  \label{fig.1} 
\end{figure}
The point cloud data handled in SLAM is often extremely large, placing a heavy burden on memory and storage. Additionally, the large data size may have a serious impact on the communication when transmitting these data (e.g., downlinking the lunar terrain map to the ground). Therefore, it is essential to compress the data size of the point clouds but also keep sufficient quality as terrain data, e.g., visual and geometrical information of the environment, which is exploited as \textit{features} in SLAM.

In this study, we propose a novel method to reduce the data size of the point cloud utilizing the Discrete Fourier Transform (DFT), as outlined in Fig. \ref{fig.1}. In the frequency-domain 2D image converted from the original 3D point cloud by DFT, the low-frequency components represent the macroscopic geometrical information of the terrain shape while the high-frequency region describes microscopic terrain features such as obstacles on the terrain. In this work, our main application scenario is the SLAM on the natural sandy terrain on the planetary surface, which doesn't have many landmark features. In this context, the proposed method preserves the low-frequency components while removing the high-frequency components in the DFT image because the values representing high-frequency components are less significant for local feature detection. In such cases, we confirmed that the error in the reconstructed point cloud after the data size reduction remains small even after removing high-frequency components. 
The main contributions of this study are as follows:
\begin{itemize}
    \item We developed the DFT-based point cloud data size reduction and evaluated the system that is effective for gently sloping terrains.
    \item We clarified the relationship between data size and map accuracy when changing the cutoff frequency, i.e., data reduction ratio.
\end{itemize}
To discuss the effectiveness and the performance of the proposed method, we applied the proposed method to the point cloud datasets including two types of terrain.

\section{RELATED WORKS}
To date, various studies have been reported on point cloud compression using height maps. In methods utilizing height maps, point clouds are projected onto a plane to generate two-dimensional height maps, and data size is reduced by applying image compression techniques.

The first example of converting point clouds to height maps for analysis was conducted by Pauly and Gross~\cite{Pauly2001}. Pauly and Gross applied DFT to point clouds and performed spectral analysis. Based on this idea, Ochotta and Saupe~\cite{Ochotta2004} proposed a method where point clouds were first partitioned using principal component analysis and then compressed via wavelet transformation. This method was later improved to further reduce compression errors using the Lloyd algorithm~\cite{Lloyd1982,Ochotta2008}. Golla and Klein applied image compression techniques such as JPEG and JPEG2000 to the generated height maps for compression~\cite{Golla2015}. In their method, point clouds were divided into several regions, and compression was performed, enabling real-time compression and decompression of point clouds at any size. Hubo et al.~\cite{Hubo2007,Hubo2008} exploited the self-similarity of height maps to compress point clouds. This method divides the surface of the point cloud into multiple patches, clusters similar patches, and replaces them with a single representative patch.

In addition, point cloud compression has been performed using octrees and neural networks~\cite{Tu2019,Cui2021,Schnabel2006,Fu2022}. While these studies on point cloud compression primarily focus on general point clouds, they do not address compression methods that take into account the geometric characteristics of the point clouds. Furthermore, the application of the Fourier transform to point clouds for the purpose of removing specific frequency components has not been explored. In this study, we propose a novel compression method that leverages the characteristics of point clouds, focusing on smooth terrains such as the lunar surface or deserts, by removing specific frequency components.

The recent work also utilized the DFT-based method by Umemura et al.~\cite{Umemura2024} for simultaneous place recognition and traversability analysis.
\section{METHOD}
\begin{figure}[t]
  \centering
  \includegraphics[width=0.4\textwidth]{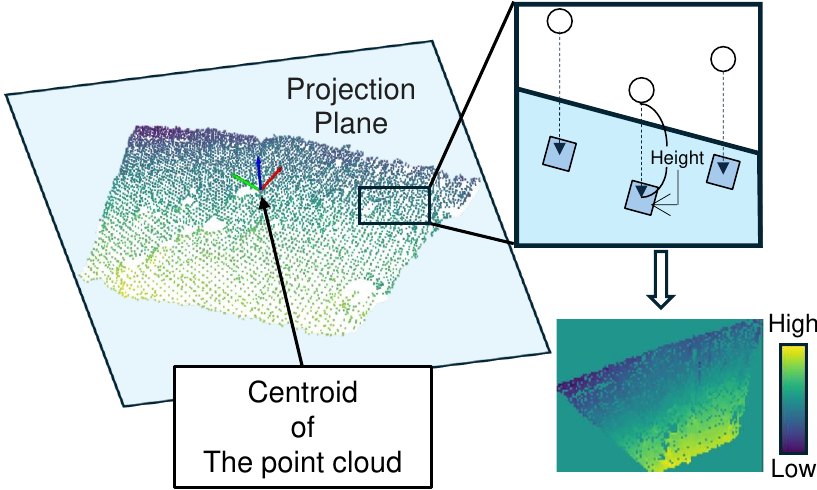} 
  \caption{Overview of DEM creation. The projection plane passes through the centroid of the point cloud.
  } 
  \label{fig:DEM} 
\end{figure}
This section explains the method for compressing point cloud maps. We assume that the exploration robot has a 3D sensor to gain the point cloud of the surrounding environment such as LiDAR, stereo camera, and Time-of-Flight camera~\cite{unoJSASS}. An overview is shown in Fig.~\ref{fig.1}. First, the original point cloud is converted into a Digital Elevation Model (DEM), which represents the three-dimensional terrain as a two-dimensional grid, with each grid point corresponding to an assigned elevation value. Next, this DEM format is processed in the Discrete Fourier Transform (DFT), which outputs the 2D image in the frequency domain. Then, the high-frequency part in the image is removed by the Low-Pass Filter (LPF). Finally, the compressed image is processed in the inverse DFT to reconstruct the point cloud data format. The following subsections detail each process.

\subsection{DEM Generation}
Our method employs a two-dimensional DFT in terms of computational efficiency. For this, the original 3D point cloud, describing the spatial terrain shapes, is represented in the 2D format, Discrete Elevation Model (DEM). In this representation, the point cloud is projected onto a predefined horizontal plane (see Fig.~\ref{fig:DEM}). Each point's height information is represented in each pixel's color. First, we define a projection plane that is parallel to the xy-plane and has a z-coordinate equal to the centroid of the point cloud and then select a plane that encloses the point cloud with the minimum bounding rectangle. When projection, the pixel size of the DEM can be arbitrarily set as the resolution. By setting a coarser resolution, the total number of points can be reduced, contributing to data reduction. This approach is similar to the uniform sampling performed by voxelization, which is commonly used in point cloud processing.

\subsection{Discrete Fourier Transform}
A two-dimensional Discrete Fourier Transform (DFT) can be applied to the DEM, which is generated as the projection of the point cloud. The DFT transforms digital signals from the spatial domain to the frequency domain. Suppose we have a two-dimensional function $f(n, m)$, where $n$ and $m$ represent spatial information with ranges of $N$ and $M$, respectively. The function $f(n, m)$ can then be transformed into a frequency representation $F(u, v)$, where $u$ and $v$ are the horizontal and vertical frequencies, respectively.
\begin{equation} F(u,v) = \sum_{n=0}^{N-1} \sum_{m=0}^{M-1} f(n,m) e^{-j 2\pi \left( \frac{un}{N} + \frac{vm}{M} \right)} \label{eq} 
\end{equation}
In equation \eqref{eq}, the dimensions of $F(u,v)$ are the same as those of $f(n,m)$.

\subsection{Low-Pass Filter}
\begin{figure}[t]
  \centering
  \includegraphics[width=0.5\linewidth]{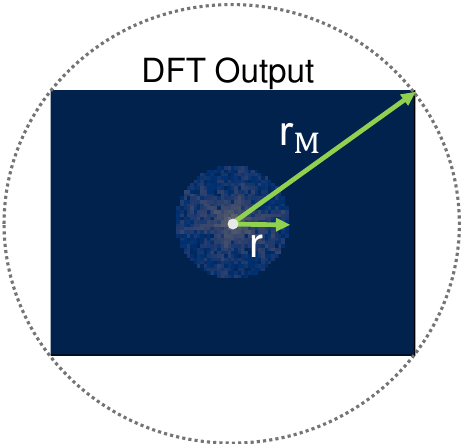} 
  \caption{Frequency-domain image as the output of DFT. $r_M$ is a fixed value for the point cloud, while $r$, the parameter to define the reduction ratio, can take arbitrary values. By varying the value of $r$, the frequency threshold for data reduction can be determined.
}
  \label{fig:cutoff} 
\end{figure}

When DFT is applied, the output arranges the low-frequency components at the center of the screen and the high-frequency components at the edges. The frequencies here represent the surface geometrical characteristics of the terrain sampled by the point cloud. The low-frequency region contributes to the overall shape of the terrain, while the high-frequency region represents finer terrain features (e.g., boulders on the surface) and noise from point cloud. In this work, we focus on gentle terrains like the Lunar or Martian sandy area, where high-frequency components are considered to have little contribution to terrain representation. Therefore, we remove the high-frequency components by means of a Low-Pass Filter (LPF).

As shown in Fig. ~\ref{fig:cutoff}, when applying an LPF, it is necessary to define a parameter that indicates which frequencies to remove. Let \(r_{\mathrm{m}}\) represent the diagonal length of the input DEM and \(r\) represent the radius for actual filtering. The cutoff ratio \(f_{\mathrm{c}}\) is defined as follows:
\begin{equation}
f_{\mathrm{c}} = 1 - \frac{r}{r_{\mathrm{M}}}
\label{eq:cutoff}
\end{equation}
\subsection{Data Compression}
In the output of the LPF, the outer region, which is filtered in a circular pattern, contains no significant values (all zeros). Therefore, we perform compression by maintaining only the values within the inner circle of the array and eliminating the redundant outer part of the array.

\subsection{Point Cloud Reconstruction}
By using the inverse Discrete Fourier Transform (IDFT), the DEM can be restored:
\begin{equation}
f(n,m) = \frac{1}{NM} \sum_{u=0}^{N-1} \sum_{v=0}^{M-1} F(u,v) e^{j 2\pi \left( \frac{un}{N} + \frac{vm}{M} \right)}
\label{eq:IDFT}
\end{equation}

Finally, performing the reverse of the DEM generation process reconstructs the point cloud from the compressed DEM.

\section{EVALUATION}
\begin{figure}[t]
  \centering
  \includegraphics[width=.9\linewidth]{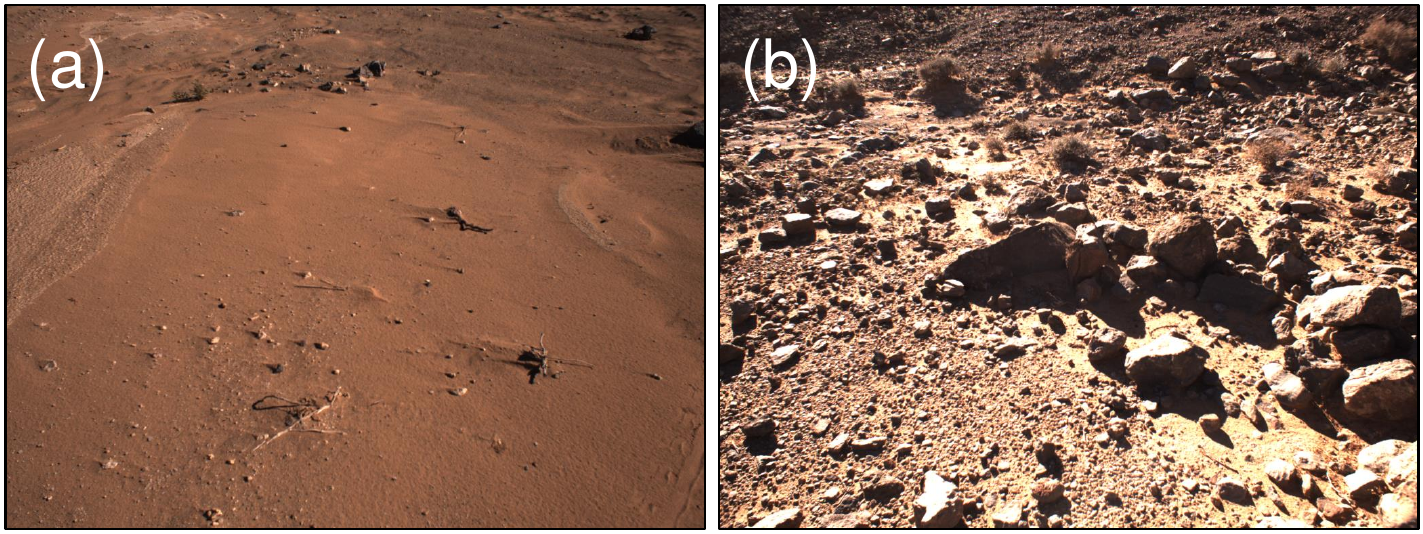} 
  \caption{Representative snaps in the MADMAX dataset.
(a) A flat sandy terrain with almost no undulations.
(b) A terrain with undulations caused by rocks.} 
  \label{fig.MADMAX} 
\end{figure}
In this study, we evaluated the proposed method by generating point clouds from stereo camera and color camera sequences of two different terrains based on a Mars-like terrain dataset, MADMAX~\cite{Meyer2021}. One terrain is mostly flat with minimal elevation changes (4707 frames), while the other terrain features significant elevation variations, including rocky areas (5948 frames). Representative frames of each terrain are shown in Fig.~\ref{fig.MADMAX}. By utilizing these different terrains, we assessed the impact of terrain relief on the system’s performance. The resolution of the generated DEMs was set to 0.1 meters.

\subsection{Evaluation Metric}
\begin{figure*}[t]
  \centering
  \includegraphics[width=.9\linewidth]{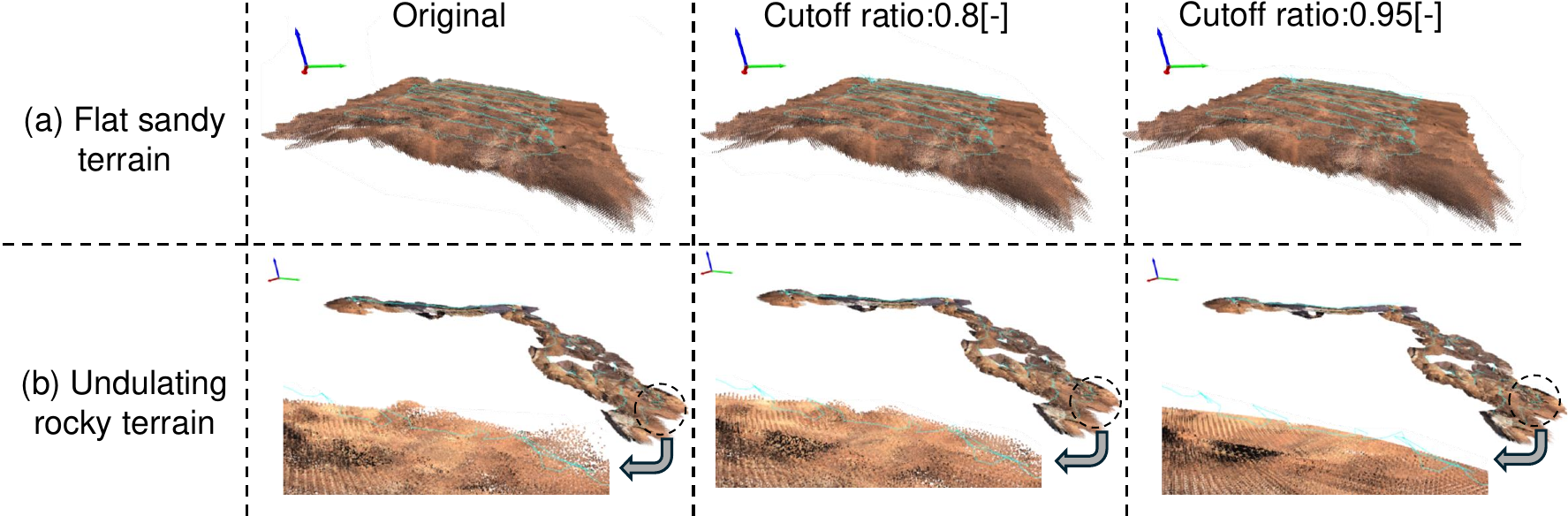} 
  \caption{
Point cloud maps and robot trajectories are generated for both terrains. From left to right: Original, Cutoff ratio $=$ 0.8, and Cutoff ratio $=$ 0.95. (b) shows an enlarged view of a specific location.
  } 
  \label{fig:map} 
\end{figure*}
To evaluate the performance of this method, the Root Mean Squared Error (RMSE), one of the widely used metric in the study of point cloud compression~\cite{Golla2015,Tu2019}, was employed. Let $N$ represent the total number of points, $\mathbf{P}$ denote the original point cloud, and $\mathbf{P}'$ the reconstructed point cloud. RMSE can be calculated using the following equations:
\begin{equation}
MSE = \frac{1}{N} \sum_{i=1}^{N} \| \mathbf{p}_i - \mathbf{p}'_i \|^2
\label{eq:MSE}
\end{equation}
\begin{equation}
RMSE = \sqrt{MSE}
\label{eq:RMSE}
\end{equation}

In other words, RMSE represents the average deviation of the reconstructed point cloud from the original point cloud. Additionally, the data reduction is evaluated using bits per point, which represents the number of bits required to represent a single point.

\subsection{Experimental results}
\begin{figure}[t]
  \centering
  \includegraphics[width=1.0\linewidth]{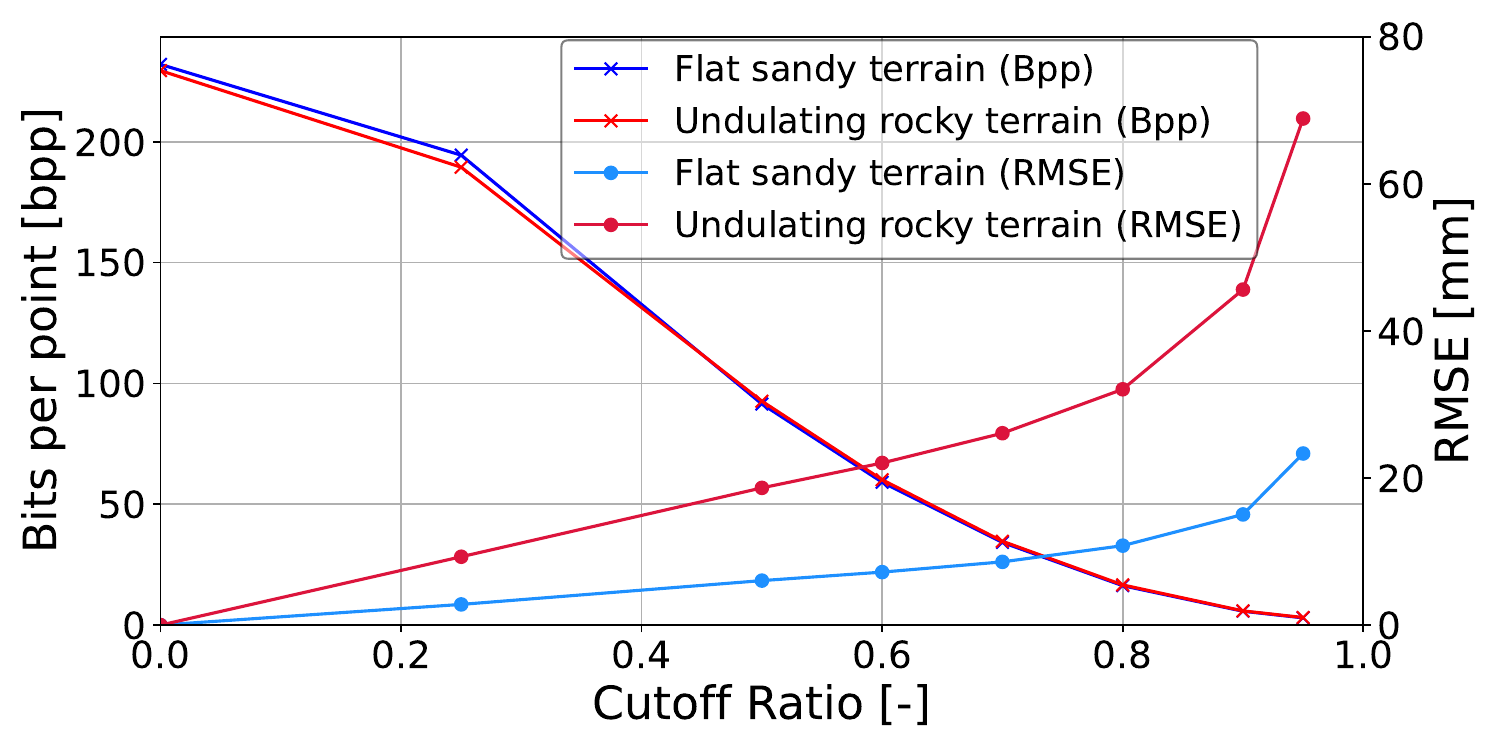} 
  \caption{Bpp and RMSE for each Cutoff ratio.
} %
  \label{fig:result} 
\end{figure}
In this subsection, a qualitative evaluation of the point clouds generated by the proposed method is first conducted, followed by a quantitative evaluation.

Fig. \ref{fig:map} shows the original point clouds and the point clouds compressed and reconstructed with Cutoff ratios of 0.8 and 0.95 for both terrains. From the figure, it can be seen that in (a) the flat sandy terrain, degradation of the point clouds due to compression is not observable. Similarly, in (b) the undulating rocky terrain, no noticeable degradation is observed in the macro view. However, in the close-up view, the difference between the original and compressed point clouds becomes more pronounced along with the cutoff ratio increases, and qualitative degradation can also be observed.

Fig. \ref{fig:result} shows the RMSE and bits per point as the cutoff ratio varies. First, it can be observed from the figure that increasing the cutoff ratio decreases the bits per point. Additionally, there is almost no difference in the reduction rate depending on the terrain, and the bits per point change similarly regardless of the terrain. Next, focusing on the relationship between the cutoff ratio and RMSE, it is confirmed that RMSE increases as the cutoff ratio increases. Furthermore, the RMSE for the undulating terrain (b) is larger than that for the flat terrain (a) at any cutoff ratio. This indicates that the proposed method enables compression with less degradation on terrains with gentle slopes. Additionally, when the cutoff ratio exceeds 0.9, RMSE sharply increases in both terrains. These results demonstrate that there is a trade-off between data size reduction and error magnitude, and selecting an appropriate cutoff ratio is crucial.

\section{CONCLUSION}
In this paper, we propose a point cloud compression method using DFT, which is suitable particularly for gently sloping terrains, exemplified in the surface on the Moon, Mars, and deserts. By focusing on the fact that high-frequency components contribute minimally to the representation of gradual terrains, we reduce the data size by filtering out these components. We also demonstrate that there is a trade-off between the cutoff ratio and the data size, showing that the proposed method is more suitable for more gradual terrains. In the experiments, we quantitatively evaluate the trade-off between the cutoff ratio and the error of the point cloud by calculating the RMSE between the reconstructed point cloud after removing specific frequency components and the original point cloud. Future challenges include verifying the real-time performance and the impact on relocalization, as well as exploring methods to select a suitable cutoff ratio for various terrains.

\end{document}